\documentclass{article}
\usepackage[accepted]{aistats2020}

\usepackage[round]{natbib}

\usepackage[utf8]{inputenc} %
\usepackage[T1]{fontenc}    %
\usepackage{hyperref}       %
\usepackage{soul} %
\usepackage{xcolor}
\usepackage{graphicx}

\definecolor{mydarkblue}{rgb}{0,0.08,0.45}

\allowdisplaybreaks

\usepackage{url}            %
\usepackage{booktabs}       %
\usepackage{amsfonts}       %
\usepackage{nicefrac}       %
\usepackage{microtype}      %

\usepackage{amsmath}
\usepackage{multirow}
\usepackage{color}
\usepackage{bm}
\usepackage{subfigure}
\usepackage{graphicx}

\usepackage{stmaryrd}
\usepackage{grffile}

\begin{document}

\twocolumn[

\aistatstitle{Regularized Autoencoders via Relaxed Injective Probability Flow}

\aistatsauthor{ Abhishek Kumar \And Ben Poole \And  Kevin Murphy}

\aistatsaddress{Google Research \And  Google Research \And Google Research } ]

\newcommand{\punt}[1]{}
\newcommand{\comment}[1]{}

\newtheorem{thm}{Theorem}[section]
\newtheorem{theorem}{Theorem}
\newtheorem{lem}{Lemma}[section]
\newtheorem{prop}{Proposition}
\newtheorem{defn}{Definition}[section]
\newtheorem{condition}{Condition}
\def\argmax{\mathop{\rm arg\,max}}
\def\argmin{\mathop{\rm arg\,min}}
\newcommand{\transpose}{{\scriptscriptstyle T}}
\newcommand{\reals}{\mathbb{R}}
\newcommand{\biggram}{\overrightarrow{\mathbf{K}}}
\newcommand{\biggramsep}{{\vK \otimes \vL}}
\newcommand{\veckernel}{\overrightarrow{k}}
\newcommand{\chvec}{\ch_{\veckernel}}
\newcommand{\chvecsep}{\ch_{k\vL}}
\newcommand{\chdict}{\ch_{\cd}}
\newcommand{\expect}{\mathbb{E}}
\newcommand{\integers}{\mathbf{Z}}
\newcommand{\naturals}{\mathbf{N}}
\newcommand{\rationals}{\mathbf{Q}}

\newcommand{\ca}{\mathcal{A}}
\newcommand{\cb}{\mathcal{B}}
\newcommand{\cc}{\mathcal{C}}
\newcommand{\cd}{\mathcal{D}}
\newcommand{\ce}{\mathcal{E}}
\newcommand{\cf}{\mathcal{F}}
\newcommand{\cg}{\mathcal{G}}
\newcommand{\ch}{\mathcal{H}}
\newcommand{\ci}{\mathcal{I}}
\newcommand{\cj}{\mathcal{J}}
\newcommand{\ck}{\mathcal{K}}
\newcommand{\cl}{\mathcal{L}}
\newcommand{\cm}{\mathcal{M}}
\newcommand{\cn}{\mathcal{N}}
\newcommand{\co}{\mathcal{O}}
\newcommand{\cp}{\mathcal{P}}
\newcommand{\cq}{\mathcal{Q}}
\newcommand{\calr}{\mathcal{R}}
\newcommand{\cs}{\mathcal{S}}
\newcommand{\ct}{\mathcal{T}}
\newcommand{\cu}{\mathcal{U}}
\newcommand{\cv}{\mathcal{V}}
\newcommand{\cw}{\mathcal{W}}
\newcommand{\cx}{\mathcal{X}}
\newcommand{\cy}{\mathcal{Y}}
\newcommand{\cz}{\mathcal{Z}}
\newcommand{\cS}{\mathcal{S}}
\newcommand{\pr}{\mathbb{P}}
\newcommand{\predsp}{\cy}  %
\newcommand{\outsp}{\cy}

\newcommand{\prxy}{P_{\cx \times \cy}}
\newcommand{\prx}{P_{\cx}}
\newcommand{\prygivenx}{P_{\cy\mid\cx}}
\newcommand{\ex}{\mathbb{E}}
\newcommand{\var}{\textrm{Var}}
\newcommand{\cov}{\textrm{Cov}}
\newcommand{\kl}{\textrm{KL}}
\newcommand{\law}{\mathcal{L}}
\newcommand{\as}{\textrm{ a.s.}}
\newcommand{\io}{\textrm{ i.o.}}
\newcommand{\ev}{\textrm{ ev.}}
\newcommand{\convd}{\stackrel{d}{\to}}
\newcommand{\eqd}{\stackrel{d}{=}}
\newcommand{\del}{\nabla}
\newcommand{\loss}{V}
\newcommand{\risk}{R}
\newcommand{\emprisk}{\hat{R}_{\ell}}
\newcommand{\lossfnl}{\risk}
\newcommand{\emplossfnl}{\emprisk}
\newcommand{\empminimizer}[1]{\hat{#1}_{\ell}} 
\newcommand{\empminimizerfa}{\hat{f}_{\ell}^{1}}
\newcommand{\empminimizerfb}{\hat{f}_{\ell}^{2}}
\newcommand{\minimizer}[1]{#1_{*}} 
\newcommand{\etal}{\textrm{et. al.}}
\newcommand{\rademacher}[1]{\calr_{#1}}
\newcommand{\emprademacher}[1]{\hat{\calr}_{#1}}

\newcommand{\trace}{\operatorname{trace}}
\newcommand{\rank}{\text{rank}}
\newcommand{\linspan}{\text{span}}
\newcommand{\spn}{\text{span}}
\newcommand{\proj}{\text{Proj}}
\def\argmax{\mathop{\rm arg\,max}}
\def\argmin{\mathop{\rm arg\,min}}

\newcommand{\bfx}{\mathbf{x}}
\newcommand{\bfy}{\mathbf{y}}
\newcommand{\bfl}{\bm{\lambda}}
\newcommand{\bfm}{\mathbf{\mu}}
\newcommand{\vmu}{\bm{\mu}}
\newcommand{\calL}{\mathcal{L}}

\newcommand{\vX}{\mathbf{X}}
\newcommand{\vY}{\mathbf{Y}}
\newcommand{\vA}{\mathbf{A}}
\newcommand{\vB}{\mathbf{B}}
\newcommand{\vE}{\mathbf{E}}
\newcommand{\vK}{\mathbf{K}}
\newcommand{\vD}{\mathbf{D}}
\newcommand{\vU}{\mathbf{U}}
\newcommand{\vL}{\mathbf{L}}
\newcommand{\vI}{\mathbf{I}}
\newcommand{\vC}{\mathbf{C}}
\newcommand{\vV}{\mathbf{V}}
\newcommand{\vM}{\mathbf{M}}
\newcommand{\vN}{\mathbf{N}}
\newcommand{\vQ}{\mathbf{Q}}
\newcommand{\vR}{\mathbf{R}}
\newcommand{\vS}{\mathbf{S}}
\newcommand{\vT}{\mathbf{T}}
\newcommand{\vZ}{\mathbf{Z}}
\newcommand{\vW}{\mathbf{W}}
\newcommand{\vH}{\mathbf{H}}
\newcommand{\vsig}{\bm{\Sigma}}
\newcommand{\vSigma}{\bm{\Sigma}}
\newcommand{\vLam}{\bm{\Lambda}}
\newcommand{\vLambda}{\bm{\Lambda}}
\newcommand{\vlam}{\bm{\lambda}}

\newcommand{\vw}{\mathbf{w}}
\newcommand{\vx}{\mathbf{x}}
\newcommand{\vxi}{\bm{\xi}}     
\newcommand{\valpha}{\bm{\alpha}}
\newcommand{\vbeta}{\bm{\beta}}
\newcommand{\veta}{\bm{\eta}}
\newcommand{\vsigma}{\bm{\sigma}}
\newcommand{\vepsilon}{\bm{\epsilon}}
\newcommand{\vdelta}{\bm{\delta}}
\newcommand{\vnu}{\bm{\nu}}
\newcommand{\vd}{\mathbf{d}}
\newcommand{\vs}{\mathbf{s}}
\newcommand{\vt}{\mathbf{t}}
\newcommand{\vh}{\mathbf{h}}
\newcommand{\ve}{\mathbf{e}}
\newcommand{\vf}{\mathbf{f}}
\newcommand{\vg}{\mathbf{g}}
\newcommand{\vz}{\mathbf{z}}
\newcommand{\vk}{\mathbf{k}}
\newcommand{\va}{\mathbf{a}}
\newcommand{\vb}{\mathbf{b}}
\newcommand{\vv}{\mathbf{v}}
\newcommand{\vr}{\mathbf{r}}
\newcommand{\vy}{\mathbf{y}}
\newcommand{\vu}{\mathbf{u}}
\newcommand{\vp}{\mathbf{p}}
\newcommand{\vq}{\mathbf{q}}
\newcommand{\vc}{\mathbf{c}}
\newcommand{\vn}{\mathbf{n}}
\newcommand{\vP}{\mathbf{P}}
\newcommand{\vG}{\mathbf{G}}

\newcommand{\hil}{\ch}
\newcommand{\rkhs}{\hil}
\newcommand{\manifold}{\cm} 
\newcommand{\cloud}{\cc} 
\newcommand{\graph}{\cg}    
\newcommand{\vertices}{\cv} 
\newcommand{\coreg}{{\scriptscriptstyle \cc}}
\newcommand{\intrinsic}{{\scriptscriptstyle \ci}}
\newcommand{\ambient}{{\scriptscriptstyle \ca}}
\newcommand{\isintrinsic}[1]{#1^{\scriptscriptstyle \ci}}
\newcommand{\isambient}[1]{#1^{\scriptscriptstyle \ca}}
\newcommand{\hilamb}{\ch^\ambient}
\newcommand{\hilintr}{\ch^\intrinsic}
\newcommand{\M}{\text{FIX ME NOW}}  %
\newcommand{\ktilde}{\tilde{k}}
\newcommand{\corkhs}{\tilde{\hil}}
\newcommand{\cok}{\ktilde}
\newcommand{\coK}{\tilde{K}}
\newcommand{\copar}{\lambda}
\newcommand{\reg}{\gamma}
\newcommand{\rega}{\gamma_{1}}
\newcommand{\regb}{\gamma_{2}}
\newcommand{\regamb}{\gamma_{\ambient}}
\newcommand{\regintr}{\gamma_{\intrinsic}}
\newcommand{\regfn}{\Omega}
\newcommand{\regfnintr}{\regfn_{\intrinsic}}
\newcommand{\regfnamb}{\regfn_{\ambient}}
\newcommand{\regfncoreg}{\regfn_{\coreg}}
\newcommand{\bq}{\begin{equation}}
\newcommand{\eq}{\end{equation}}
\newcommand{\ba}{\begin{eqnarray}}
\newcommand{\ea}{\end{eqnarray}}

\newcommand{\spana}{\cl^{1}}
\newcommand{\spanb}{\cl^{2}}
\newcommand{\ha}{\hil^{1}}
\newcommand{\hb}{\hil^{2}}
\newcommand{\fa}{f^{1}}
\newcommand{\fb}{f^{2}}
\newcommand{\Fa}{\cf^{1}}
\newcommand{\Fb}{\cf^{2}}
\newcommand{\ka}{k^{1}}
\newcommand{\kb}{k^{2}}
\newcommand{\vcok}{\boldsymbol{\cok}}
\newcommand{\vka}{\vk^{1}}
\newcommand{\vkb}{\vk^{2}}
\newcommand{\Ka}{K^{1}}
\newcommand{\Kb}{K^{2}}
\newcommand{\kux}{\vk_{Ux}}
\newcommand{\kuz}{\vk_{Uz}}
\newcommand{\kuu}{K_{UU}}
\newcommand{\kul}{K_{UL}}
\newcommand{\klu}{K_{LU}}
\newcommand{\kll}{K_{LL}}
\newcommand{\kuxa}{\vk_{Ux}^{1}}
\newcommand{\kuza}{\vk_{Uz}^{1}}
\newcommand{\Kuua}{K_{UU}^{1}}
\newcommand{\Kula}{K_{UL}^{1}}
\newcommand{\Klua}{K_{LU}^{1}}
\newcommand{\Klla}{K_{LL}^{1}}
\newcommand{\kuxb}{\vk_{Ux}^{2}}
\newcommand{\kuzb}{\vk_{Uz}^{2}}
\newcommand{\Kuub}{K_{UU}^{2}}
\newcommand{\Kulb}{K_{UL}^{2}}
\newcommand{\Klub}{K_{LU}^{2}}
\newcommand{\Kllb}{K_{LL}^{2}}

\newcommand{\Ksum}{S}
\newcommand{\ksum}{s}
\newcommand{\vksum}{\vs}

\newcommand{\intrinsicRegMat}{M_{\intrinsic}}
\newcommand{\coregPointCloudMat}{M_{\coreg}}
\newcommand{\vid}[1]{#1^\text{vid}}
\newcommand{\aud}[1]{#1^\text{aud}}
\newcommand{\bad}[1]{#1_\text{bad}}
\newcommand{\empcompat}{\hat{\chi}}

\newcommand{\nn}{\ensuremath{k}}

\newcommand{\uva}{\ushort{\va}}
\newcommand{\uvf}{\ushort{\vf}}
\newcommand{\uvg}{\ushort{\vg}}
\newcommand{\uvk}{\ushort{\vk}}
\newcommand{\uvw}{\ushort{\vw}}
\newcommand{\uvh}{\ushort{\vh}}
\newcommand{\uvbeta}{\ushort{\vbeta}}
\newcommand{\uA}{\ushort{A}}
\newcommand{\uG}{\ushort{G}}

\def\la{{\langle}}
\def\ra{{\rangle}}
\def\R{{\reals}}

\newcommand{\mbf}[1]{\mathbf{#1}}
\newcommand{\mbb}[1]{\mathbb{#1}}
\newcommand{\mcal}[1]{\mathcal{#1}}
\newcommand{\remove}[1]{}
\newcommand{\nuc}[1]{\left\lVert #1\right\rVert_*}
\newcommand{\spec}[1]{\left\lVert #1\right\rVert_2}
\newcommand{\frob}[1]{\left\lVert #1\right\rVert_F}
\newcommand{\norm}[1]{\left\lVert #1\right\rVert}
\newcommand{\abs}[1]{\left\lvert #1\right\rvert}
\newcommand{\obs}[1]{P_\Omega({#1})}

\newcommand{\red}[1]{{\color{red} #1}}

\newcommand{\ball}{\mathcal{B}}
\newcommand{\sphere}{\mathcal{S}}
\newcommand{\X}{\mathcal{X}}
\newcommand{\domain}{\mathcal{C}}
\newcommand{\polytope}{\mathcal{P}}
\newcommand{\Am}{\bm{A}}
\newcommand{\Bm}{\bm{B}}
\newcommand{\xv}{\bm{x}}
\newcommand{\yv}{\bm{y}}
\newcommand{\zv}{\bm{z}}
\newcommand{\sv}{\bm{s}}
\newcommand{\av}{\bm{a}}
\newcommand{\bv}{\bm{b}}
\newcommand{\pv}{\bm{p}}
\newcommand{\uv}{\bm{u}}
\newcommand{\wv}{\bm{w}}
\newcommand{\omegav}{\bm{\omega}}
\newcommand{\mv}{\bm{m}}
\newcommand{\GW}{G_\setC}%
\newcommand{\G}{G}
\newcommand{\0}{\bm{0}}
\newcommand{\id}{\bm{\iota}}
\newcommand{\alphav}{\bm{\alpha}}
\newcommand{\nuv}{\bm{\nu}}
\newcommand{\thetav}{\bm{\theta}}
\newcommand{\lambdav}{\bm{\lambda}}
\newcommand{\epsilonv}{\bm{\epsilon}}
\newcommand{\row}{\text{row}}
\newcommand{\col}{\text{col}}
\newcommand{\lft}{\text{left}}
\newcommand{\rgt}{\text{right}}
\newcommand{\one}{\mathbf{1}} %

\newcommand{\todo}[1]{\marginpar[\hspace*{4.5em}\textbf{TODO}\hspace*{-4.5em}]{\textbf{TODO}}\textbf{TODO:} #1}

\newcommand{\aatop}[2]{\genfrac{}{}{0pt}{}{#1}{#2}}
\newcommand{\ie}{\textit{i}.\textit{e}.}
\newcommand{\eg}{\textit{e}.\textit{g}.}

\newcommand{\blue}[1]{\textcolor{blue}{#1}}

\newcommand{\proposed}{InjFlow}
\newcommand{\proposedln}{InjFlow$^{\ell n}$}

\begin{abstract}
Invertible flow-based generative models are an effective method for learning to generate samples, while allowing for tractable likelihood computation and inference. 
However, the invertibility requirement restricts models to have the same latent dimensionality as the inputs.
 This imposes significant architectural, memory, and computational costs,  making them more challenging to scale than other classes of generative models such as Variational Autoencoders (VAEs). 
 We propose a generative model based on probability flows that does away with the bijectivity requirement on the model and only assumes injectivity. 
This also provides another perspective on regularized autoencoders (RAEs), with our final objectives resembling RAEs with specific regularizers that are derived by lower bounding the probability flow objective. 
We empirically demonstrate the promise of the proposed model, improving over VAEs and AEs in terms of sample quality. 
\end{abstract}

\section{Introduction}
\label{sec:intro}
Invertible flow-based generative models \citep{dinh2016density,kingma2018glow} have recently gained traction due to several desirable properties: (i) exact log-likelihood calculation (unlike VAEs that maximize a lower bound \citep{kingma2013auto, rezende2014stochastic}), (ii) exact inference of latent variables, and (iii) good sample quality relative to VAEs. %

However, a limitation of invertible flow models is that they require invertibility on the full ambient space, resulting in a latent space with the same dimensionality as the input data. This requirement leads to larger models with higher memory and computational costs that are more difficult to scale than VAE and GAN counterparts that have lower-dimensional latent spaces \citep{goodfellow2014generative, kingma2018glow}. This lack of dimensionality reduction also makes it difficult to capture high-level generative factors directly in \emph{individual} latent dimensions, 
a property that is often 
argued to be desirable for generative models \citep{higgins2016beta,narayanaswamy2017learning,kumar2017variational,kim2018disentangling,chen2018isolating}. 

In this work we propose a generative model of data based on probability flows that relaxes the bijectivity requirement.  
The model $g$ maps low dimensional latents $z\in Z=\reals^d$ to samples in the image of $g$, residing in the much higher-dimensional ambient space $\reals^D$. A probability distribution in the latent space (\eg, standard normal) is pushed forward by the mapping $g$ to induce a distribution on the \emph{image}, $g(Z)$. By taking the mapping $g$ to be one-to-one or injective and differentiable, we can use a change of variables theorem to obtain a closed form for the distribution over $g(Z)$. While the resulting log-likelihood in ambient space is intractable, we can form tractable lower-bounds using stochastic approximations to obtain  objectives amenable to stochastic first-order optimization. Relaxing the bijectivity requirement loses the ability to provide exact likelihoods %
for data points that lie off the image $g(Z)$. %
In this work, we limit ourselves to using the derived flow-based objective for learning a sampling mechanism that always generates samples from the image $g(Z)$. This is in contrast to VAEs where generated samples lie off the image $g(Z)$ due to the presence of an additional distribution at the output of the decoder.\footnote{One can augment the decoder with an ambient noise distribution (\eg, Gaussian) either as part of the training objective or post-hoc after training \citep{wu2017quantitative} for estimating log-likelihoods, but we do not consider that here.}

\begin{figure*}[!htbp]
	\centering
	\includegraphics[width=0.43\textwidth]{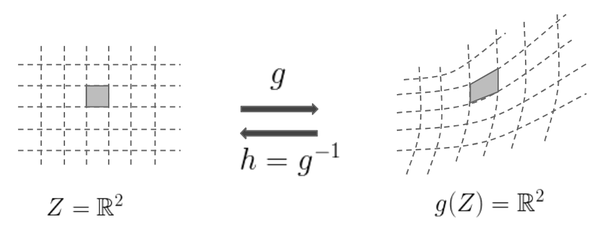}
	\qquad\qquad
	\includegraphics[width=0.47\textwidth]{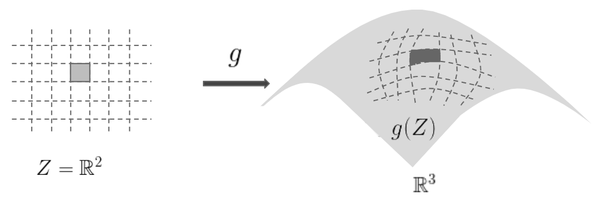}
	\caption{Schematic of invertible (left) vs. injective (right) mappings. Invertible flow models require that $Z$ and $X$ be the same dimensionality, and the mapping $g$ to be invertible on the full domain. In contrast, injective mappings can have lower-dimensional $Z$ but are invertible only on the image of $g$ (shaded).}
	\label{fig:invinj}
\end{figure*}
Our final objective, although derived from the probability flow perspective, resembles a regularized autoencoder with an additional prior log probability term and an annealing of the weight on the reconstruction loss that increases over time. This flow perspective  motivates several commonly used autoencoder regularization strategies, \eg, those in \citet{ghosh2019variational}. %
We evaluate the relaxed injective flow models on MNIST, CIFAR-10, and CelebA, where we observe better FID scores compared to VAEs, and \citet{ghosh2019variational}. Our 
results demonstrate that these models provide an efficient and tractable mechanism for training neural samplers with compressed latent spaces.

\section{Formulation}
\label{sec:form}

\subsection{Invertible Flows}
Let $g:Z\to X$ be a generator mapping from latents to data, assumed to be differentiable everywhere. If $g$ is a bijection with $h=g^{-1}:X\to Z$, then $X$ and $Z$ must have the same dimensionality and we can write the distribution induced over $X$ in terms of the distribution over $Z$ using change of variables formula:
\begin{align}
\begin{split}
\ln p_x(x) &= \ln p_z(h(x)) + \ln \left| \det J_h(x)\right| \\
&= \ln p_z(h(x)) - \ln \left| \det J_g(h(x))\right|
\end{split}
\label{eq:inv}
\end{align}
\noindent where $J_h(x)$ and $J_g(z)$ are the Jacobians of $h$ and $g$ at $x$ and $z$, respectively. Invertible flow models \citep{dinh2016density,papamakarios2017masked,kingma2018glow} optimize \eqref{eq:inv} to learn a generative model of the data. They provide tractable objectives by structuring the generator so that the inverse and the log-det-Jacobian terms are tractable. Recent work on invertible residual nets  \citep{behrmann2018invertible} makes use of certain approximations to get a tractable objective for invertible flows with ResNets having Lipschitz constrained residual blocks. More recently, \citet{behrmann2020on} studied the numerical stability of invertible flow models, finding that, in practice, numerical issues may prevent the models from being invertible in certain regions even slightly off the data manifold. 

\subsection{Injective Flows}
\remove{
\begin{figure}[t]
	\centering
	\includegraphics[width=0.7\textwidth]{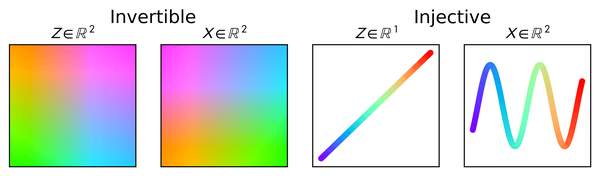} 
	\caption{Schematic of invertible (left) vs. injective (right) functions, colored by the value of the latent $z$. Invertible flow models require that $Z$ and $X$ be the same dimensionality, and the mapping $g$ to be invertible on the full domain. In contrast, injective mappings can have lower-dimensional $Z$ but are invertible only on the image of $g$.}
	\label{fig:invinj}
\end{figure}
}
We are interested in developing probability flow-based models for the setting when the dimensionality of the latent space $Z$ is much lower than the data dimensionality, i.e., $Z=\reals^d$ and $X=\reals^D$, where $D\gg d$. %
We can obtain a change of variables formula for this setting by looking at how an infinitesimal volume element $dz$ at $z\in Z$ is transformed by the mapping $g$. The theory of integration on manifolds tells us that the mapping $g$ transforms an infinitesimal volume element $dz$ at $z\in Z$ to a corresponding volume  $[\det(J_g(z)^\top J_g(z))]^{1/2} dz$ on $g(Z)$  \citep{boothby1986introduction}, where $g(Z)=\{g(z):z\in Z\}$ is the image of $Z$ under $g$. 
If we assume $g$ is an injective function and thus invertible when seen as a mapping $g:Z\to g(Z)$, we can write the \emph{probability flow} from $Z$ to $g(Z)$ as:
\begin{align}
\begin{split}
\ln p_x(x) = \ln p_z(z) - \frac{1}{2} \ln & \left| \det J_g(z)^\top J_g(z)\right|, \\
& \text{ s.t. } x=g(z). 
\end{split}
\label{eq:inj}
\end{align}
Note that $\ln |\det J_g(z)^\top J_g(z)|=\ln \det J_g(z)^\top J_g(z)$ as $J_g(z)^\top J_g(z)$ is a positive definite matrix.  Figure~\ref{fig:invinj} presents a schematic of invertible and injective functions transforming an infinitesimal volume element. The more familiar change of variables formula in \eqref{eq:inv} can be derived as the special case when $J$ is square and thus $\det J^TJ=(\det J)^2$.
In order to avoid solving an inverse problem for every $x$ in our data (\ie, finding $z$ for every $x$ s.t. $x=g(z)$), we assume the existence of an encoder $h:X\to Z$ such that $g(h(x))=x$ for every $x\in g(Z)$. This lets us write
\begin{align}
\begin{split}
\ln p_x(x) = \ln p_z(h(x)) - \frac{1}{2} \ln & \left| \det J_g(h(x))^\top J_g(h(x))\right|, \\
& \text{ s.t. } x=g(h(x)). 
\end{split}
\label{eq:injenc}
\end{align}
 Optimizing the $\ln \left| \det J_g(h(x))^\top J_g(h(x))\right|$ term exactly may be computationally challenging for large models as it requires computing a $D \times d$ Jacobian matrix for every data point (however, it could still be tractable for smaller models where the latent dimensionality $d$ is small). 
 We propose two ways to lower bound the log likelihood in \eqref{eq:injenc} in order to obtain a tractable objective we can maximize.
 Let the singular values of $J_g(h(x))$ be given by  $\{s_i\}_{i=1}^d$. Using the inequality $\ln x \le \frac{x}{\lambda} + \ln \lambda - 1$, for all $x>0$ and $\lambda>0$ (based on concavity of the log), we have
 \begin{align*}
\ln p(x) &= \ln p(h(x)) - \frac{1}{2} \ln \det [J_g(h(x))^\top J_g(h(x))] \\
&= \ln p(h(x)) - \frac{1}{2} \sum \ln s_i^2 \\
&\geq\,\,\, \ln p(h(x)) - \frac{1}{2} \sum_i \left(\frac{s_i^2}{\lambda}+\ln\lambda -1\right) \\
= \ln \,& p(h(x)) - \frac{1}{2\lambda}  \left\lVert J_g(h(x))\right\rVert_F^2 - \frac{d}{2}\ln\lambda + \frac{d}{2}, \stepcounter{equation}\tag{\theequation}\label{eq:injlower2}\\
& \qquad\qquad\qquad\qquad\qquad \text{ s.t. } x=g(h(x)). 
\end{align*}

This lower bound is maximized for $\lambda=\frac{1}{d}\sum_i s_i^2$. Substituting it into \eqref{eq:injlower2}, we get 

\begin{align*}
\ln p(x) & \geq\,\, \ln p(h(x)) - \frac{d}{2} \ln \left( \frac{1}{d} \sum_i s_i^2\right) \\
&= \ln p(h(x)) - \frac{d}{2} \ln \left(\frac{1}{d} \left\lVert J_g(h(x))\right\rVert_F^2\right), \stepcounter{equation}\tag{\theequation}\label{eq:injlower}\\
& \qquad\qquad\qquad\qquad\qquad \text{ s.t. } x=g(h(x)). 
\end{align*}
The bounds in \eqref{eq:injlower2} and \eqref{eq:injlower} are also computationally expensive but we will show how to form efficient stochastic approximations in the next section.

{\bf Tightness of the bounds.~} 
The inequality in \eqref{eq:injlower2} is tight when all singular values are equal to $\lambda$. %
Note that it is also possible to use separate $\lambda_i$ corresponding to each $s_i$ and tune these as hyperparameters to improve upon the tightness of the bound, however we do not explore this for the sake of simplicity. We will still tune the hyperparameter $\lambda$ in \eqref{eq:injlower2} to see how it performs against the objective in \eqref{eq:injlower}. 

The objectives in \eqref{eq:injlower} and \eqref{eq:injlower2} are constrained optimization problems that can be solved with a variety of approaches. Recent work on VAEs has used the augmented Lagrangian method to enforce reconstruction constraints \citep{rezende2018taming}, but here we use the penalty method for its simplicity \citep{bertsekasnonlinear}. Applying penalty method to \eqref{eq:injlower}, we get:
\begin{align}
\begin{split}
\ln p_z(h(x)) - \frac{d}{2} \ln \left\llbracket\frac{1}{d}\left\lVert J_g(h(x))\right\rVert_F^2\right\rrbracket - \mu \lVert x-g(h(x))\rVert_2^2,
\end{split}
\label{eq:injpen}
\end{align}
\noindent where $\mu$ is a positive real that is increased as the optimization progresses \citep{bertsekasnonlinear}. %
Optimizing \eqref{eq:injpen} can still be computationally demanding as it involves computing the full Jacobian of the generator. %
We can use Hutchinson's trace estimator \citep{hutchinson1990stochastic} to avoid explicitly materializing the full Jacobian. Hutchinson's trace estimator is based on the fact that $\text{tr}(A)=\expect_v \text{tr}(Avv^\top)=\expect_v v^\top Av$ for any random vector $v$ s.t. $\expect_v vv^\top = I$. We write the Frobenius norms of the Jacobian %
as $\lVert J_g(h(x))\rVert_F^2 %
= \text{tr}(J_g(h(x))^\top J_g(h(x))) = \expect_v v^\top J_g(h(x))^\top J_g(h(x)) v = \expect_v \lVert J_g(h(x)) v\rVert_2^2$ for $v\sim N(0,I_d)$. 
We further employ the unbiased Monte-Carlo estimation $\expect_v \lVert J_g(h(x)) v\rVert_2^2\approx \frac{1}{k}\sum_{i=1}^k \lVert J_g(h(x)) v_i\rVert_2^2$, and use one Monte-Carlo sample per example ($k=1$) in a minibatch.
This leads to an unbiased estimator of the bound when used with objective \eqref{eq:injlower2}, and the expectation of the Monte-Carlo approximation \emph{remains a lower bound} on the log likelihood. 

When used with the objective in \eqref{eq:injlower}, Hutchinson's estimator leads to a biased estimator of $\ln\lVert J_g(h(x))\rVert_F^2$, as 
the expectation of this estimator is smaller than $\ln\expect_v \lVert J_g(h(x)) v\rVert_2^2$. This results in an estimate whose expectation may \emph{no longer be a bound on the log likelihood}. Similar issue arises in earlier works that try to do Monte Carlo estimation for $\ln \expect_x f(x)$ \citep{li2016renyi,rhodes2018variational}.
In spite of no longer bounding log-likelihood,
we find that this approximation is still effective in practice to train neural samplers. %
Using this Monte-Carlo approximation yields
\begin{align}
\begin{split}
\ln p_z(h(x)) - \frac{d}{2} \ln \lVert J_g(h(x)) v\rVert_2^2 - \mu \lVert x-g(h(x))\rVert_2^2, 
\end{split}
\label{eq:injpenst}
\end{align}
\noindent with $v\sim N(0,I_d)$ and ignoring the constant terms (the factor of $\frac{1}{d}$ can also be absorbed in $v$).  %
We use automatic differentiation\footnote{If the automatic differentiation framework only allows for reverse mode AD, one can use $\expect_v \lVert J_g(h(x))^\top v\rVert_2^2$ with $v\sim N(0,I_D)$ for $\lVert J_g(h(x))\rVert_F^2$, instead of $\expect_v \lVert J_g(h(x)) v\rVert_2^2$ with $v\sim N(0,I_d)$.} to optimize the term containing the Jacobian-vector product. However, we observe numerical instabilities while training models for some configurations (CIFAR-10 with $p_z=N(0,I)$). 
In these cases, we use the finite difference approximation:  %
\begin{align}
\lVert J_g(h(x)) v\rVert_2^2 \approx \frac{\lVert g(z+\epsilon v)-g(z)\rVert^2}{\epsilon^2},
\end{align}
\noindent with small $\epsilon\, (=0.01)$, and $v\sim N(0,I_d)$. 

For $g$ to be injective, a necessary condition is to constrain all singular values of $J_g(h(x))$ to be bounded away from zero. 
Instead of directly enforcing this which can be computationally challenging, we simply enforce that $\lVert J_g(h(x)) v\rVert_2$ is greater than a threshold $\eta$ for all $v$ with $\lVert v\rVert=1$. A similar approach was used by \citet{odena2018generator}. 
There are scenarios where positivity of singular values does not ensure global injectivity, \ie, there may exist $z_1, z_2$ s.t. $x=g(z_1)=g(z_2)$ (see self-intersections in \citet{lagrange2007sufficient}) . Suppose $h(x)=z_1$ in this case, then the lower bounds in \eqref{eq:injlower} and \eqref{eq:injlower2} are still valid since $\ln p(x) \geq \ln p(h(x)) - \frac{1}{2} \ln \det [J_g(h(x))^\top J_g(h(x))]$. 

While training, we take the latent space distribution $p_z$ to be an isotropic Gaussian distribution $N(0,\sigma^2 I_d)$, which reduces the first term in bounds \eqref{eq:injlower2} and \eqref{eq:injlower} to be $\frac{1}{2\sigma^2}\lVert h(x)\rVert_2^2$. 
Our final minimization objective corresponding to the lower bound of \eqref{eq:injlower} is given by
\begin{align}
\begin{split}
\min_{h,g}\,\, \expect_{x,v}\bigg(\,\, & \frac{1}{2\sigma^2}\lVert h(x)\rVert_2^2 + \mu \lVert x-g(h(x))\rVert_2^2 \,\, + \\ & \frac{d}{2} \ln \left\llbracket \max(\lVert J_g(h(x)) v\rVert_2^2, \eta^2\lVert v\rVert^2)\right\rrbracket \,+ \\
 &  \mu_{in}  \left[\frac{1}{\lVert v\rVert}\lVert J_g(h(x)) v\rVert_2-\eta\right]_{-}^2
\bigg)
\end{split}
\label{eq:final}
\end{align}
\noindent where $v\sim N(0,I_d)$, $[a]_{-}=\min(a,0)$, and $\mu_{in}$ is a positive penalty on the constraint enforcing the local injectivity of the generator. Both $\mu$ and $\mu_{in}$ are increased over the course of optimization \citep{bertsekasnonlinear}. %

Following similar steps of forming an unconstrained objective using the penalty method and using Monte-Carlo estimation for $\frob{J_g(h(x))}^2$, we obtain the following minimization objective corresponding to the lower bound of \eqref{eq:injlower2}:
\begin{align}
\begin{split}
\min_{h,g}\,\, & \expect_{x,v}\bigg(\,\, \frac{1}{2\sigma^2}\lVert h(x)\rVert_2^2 + \mu \lVert x-g(h(x))\rVert_2^2 \,\, + \\ & \frac{1}{2\lambda} \left\llbracket \max(\lVert J_g(h(x)) v\rVert_2^2, \eta^2\lVert v\rVert^2)\right\rrbracket \,+ \\
 & \mu_{in}  \left[\frac{1}{\lVert v\rVert}\lVert J_g(h(x)) v\rVert_2-\eta\right]_{-}^2\bigg),
\end{split}
\label{eq:final2}
\end{align}
\noindent where $\lambda$ is a fixed hyperparameter (which is not optimized over but can be tuned as discussed earlier). 
We optimize the objectives \eqref{eq:final} and \eqref{eq:final2} with respect to parameters of both the generator $g$ and the encoder $h$. 

\subsubsection{Sampling from the model}
Although the injective flow model transforms an isotropic Gaussian prior to the data distribution, in practice we observe that the distribution of encoded data points (``aggregate posterior'') deviates from the prior distribution, which is also reflected in poor quality of generated samples. 
Note that this is not linked to invertibility and can happen even when the network is perfectly invertible. This issue is not specific to our model and is present even in VAEs and bijective flow based models. For VAEs, recent work highlighted this issue in the case of modeling a data distribution that lies along a low-dimensional manifold (Sec. 4 in \citet{dai2019diagnosing}) and proposed fitting another distribution %
on the encoded latents after training the VAE. For invertible-flow models the Euclidean norm of the latent codes is often different from the typical set of the prior, indicating a systematic aggregate posterior-prior mismatch (see \citet{waic}, and Fig.~8 in \citet{kingma2018glow}). 

Hence, for sampling from the model, we fit a distribution over the encoded training data in the latent space after the model has been trained, an approach taken by several recent works \citep{van2017neural,dai2019diagnosing,ghosh2019variational}. \citet{dai2019diagnosing} train another VAE on the encoded training data to get a complex post-fit prior. However, in this paper we experiment with fitting a Gaussian prior and a mixture of 10 Gaussians similar to \citet{ghosh2019variational}.

\comment{
{\bf Remarks.~} \\
\noindent {\bf 1.} Monte Carlo estimation used for the term $\ln \expect_v \lVert J_g(h(x))^\top J_g(h(x))\rVert^2$ in \eqref{eq:injpenst} results in a biased estimator. The expectation of this estimator is lower than $\ln \expect_v \lVert J_g(h(x))^\top J_g(h(x))\rVert^2$ which may not result in a lower bound on the log liklihood. Similar issue arises in some earlier works that try to do Monte Carlo estimation for $\ln \expect_x f(x)$ \citep{li2016renyi,rhodes2018variational}. In practice, we do not find it to be an  issue. 

\noindent {\bf 2.} The regularizer in Eq. \eqref{eq:injenc}, which is equivalent to $\sum \ln s_i^2$, minimizes the sum of log singular values of the Jacobian $J_g(h(x))$ at the encoded training points $x$. This encourages the mapping $g$ to be smooth at the training points. However the gradient is dominated by the smallest singular values. In \eqref{eq:injlower}, we upper bound this term by $\frac{1}{d}\ln \frac{1}{d}\sum_i s_i^2$ where the gradient is dominated by the largest singular value and tries to push all singular values towards zero.
}
\section{Related Work}
\label{sec:related}
Our work is similar in spirit to the recent work of \citet{ghosh2019variational, van2017neural, dai2019diagnosing},  that find regularized autoencoders paired with a learned prior produces high-quality samples.
Our work provides another perspective on the regularized autoencoder (RAE) objective in \citep{ghosh2019variational}, wherein the regularization terms arise naturally from approximating the log-likelihood objective of the injective probability flow. 
\citet{ghosh2019variational} motivate the RAE objective by considering constant posterior-variance VAEs and connecting stochasticity at the decoder's input (arising by sampling from $q(z|x)$) to  \emph{smoothness} of the decoder. Recently, \citet{kumar2020implicit} analyzed the implicit regularization in $\beta$-VAEs, deriving a regularizer that also depends on the Jacobian of the deocder but has a different form. 

Regularized autoencoders have been widely studied in earlier works as well \citep{rifai2011contractive, alain2014regularized, poole2014analyzing}. Contractive autoencoders \citep{rifai2011contractive} also penalize the Frobenius norm of the Jacobian, however the penalty is on the {\em encoder} Jacobian, which is different from our penalty on the {\em decoder} Jacobian. Most of these prior works on RAEs has been on improving the quality of the encoder for downstream tasks, wheres we are primarily interested in the quality of the generator for producing samples. Recent work has turned to regularizing autoencoders for sample quality as well, for example improving interpolation quality using an adversarial training objective \citep{berthelot2018understanding}.

\citet{krusinga2019understanding} recently used Eq.~\eqref{eq:inj} to get density estimates for trained GANs. However as we noted earlier, these density estimates are by nature undefined for unseen real examples which may lie off the manifold.

Several earlier works have also used spectral regularizers in training generative models. \citet{miyato2018spectral} encourage Lipschitz smoothness of the GAN discriminator by normalizing the spectral norm of each layer. \citet{odena2018generator} study the spectral properties of the Jacobian of the generator and its correlation with the quality of generated samples. They empirically observe that regularizing the condition number of the Jacobian leads to more stable training and improved generative model.

\vspace{-3mm}
\section{Experiments}
\label{sec:exp}
\vspace{-2mm}
{\bf Datasets.~}
Our experimental framework is based on \citet{ghosh2019variational}. 
We evaluate our proposed model and baselines on three publicly available datasets: CelebA \citep{liu2015deep}, CIFAR-10 \citep{krizhevsky2009learning}, and MNIST \citep{Lecun98gradient-basedlearning}.
We use $64\times 64$ cropped images for CelebA faces as used in several prior works. Image size for CIFAR-10 and MNIST is $32\times 32$ and $28\times 28$, respectively. \\

\begin{table*}[t]
	\centering
	\caption{FID scores (lower is better). \emph{Rec:} score for reconstructed test data, $\mcal{N}:$ score for decoded samples from a Gaussian prior with full covariance fit to encoded training samples, \emph{GMM:} score for decoded samples from a GMM fit to encoded training samples. \proposed~and \proposedln~are the models obtained from the objectives \eqref{eq:final2} and \eqref{eq:final}, respectively (with superscript ${\ell n}$ denoting the presence of $\log$ with the Frobenius term in \eqref{eq:injlower} and \eqref{eq:final}.}
	\vspace{2mm}
	\begin{tabular}{c c c c c c c c c c c c}
		\toprule[0.25ex]
		& \multicolumn{3}{c}{MNIST} && \multicolumn{3}{c}{CIFAR10}&& \multicolumn{3}{c}{CelebA} \\ 
		\cmidrule{2-4}\cmidrule{6-8}\cmidrule{10-12} 
		& \multirow{3}{*}{Rec.} & \multicolumn{2}{c}{Samples} && \multirow{3}{*}{Rec.} & \multicolumn{2}{c}{Samples} && \multirow{3}{*}{Rec.} & \multicolumn{2}{c}{Samples} \\
		\cmidrule{3-4} \cmidrule{7-8}  \cmidrule{11-12}
		& & $\mcal{N}$ & GMM && & $\mcal{N}$ & GMM && & $\mcal{N}$ & GMM\\
		\cmidrule{2-4}\cmidrule{6-8}\cmidrule{10-12}
VAE & 65.10 & 57.04 & 62.08 && 176.5 & 169.1 & 184.3 && 62.36& 72.48 & 67.82\\
$\beta$-VAE & 7.91 & {\bf 24.31} & {\bf 8.12} && 43.86 & 83.59 & 71.56 && 30.06 & 50.66 & 42.77 \\
AE & 8.69 & 43.40 & 12.14 &&41.45 & 81.13& 70.97 && 30.16 &51.48 & 43.49  \\
CAE & 10.51 & 45.18 & 12.90 && 41.13 & 81.53 & 70.11 && 31.12 & 48.13 & 40.67\\
AE+L2 & 7.76 & 34.27 & 9.69 && 43.02 & 81.28 & 70.13 && 29.97 & 50.02 & 42.09\\
AE+SN & 8.07 & 37.19 & 11.84 && 41.34 & 81.35 & 70.94 && 31.21 & 51.13 & 43.33\\
\proposedln & 7.40 & 35.96 & 9.93 && {\bf 40.11} &{\bf 78.78}&{\bf 68.26}&& {\bf 27.93} & {\bf 47.70} & {\bf 40.23} \\
\proposed & {\bf 6.0} & 42.65 & 11.43 && 40.86 & 79.67 & 68.37 && 28.51 & 49.01 &  40.57\\
		\bottomrule[0.25ex]
	\end{tabular}
	\label{tab:results}
\end{table*}
\begin{figure*}[h]
	\centering
	\includegraphics[width=\textwidth]{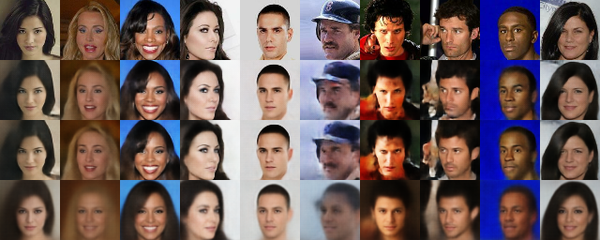} 
	\caption{CelebA reconstructions: \emph{Top to bottom:} Randomly sampled test examples, \proposed~reconstructions, Autoencoder reconstructions, VAE reconstructions.}
	\label{fig:recon_celeba}
\end{figure*}

\begin{figure*}[!htbp]
	\centering
	\includegraphics[width=0.42\textwidth]{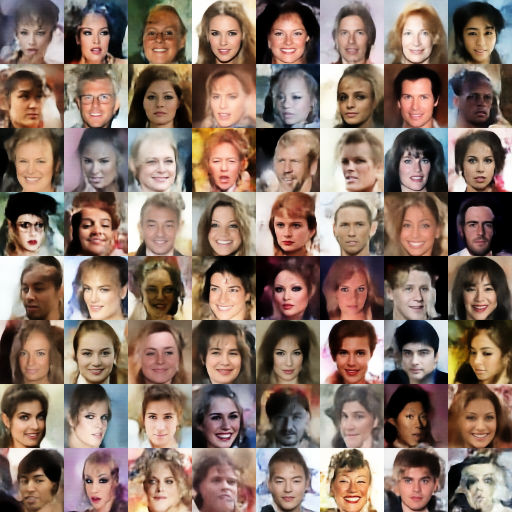} \hspace{2mm}
	\includegraphics[width=0.42\textwidth]{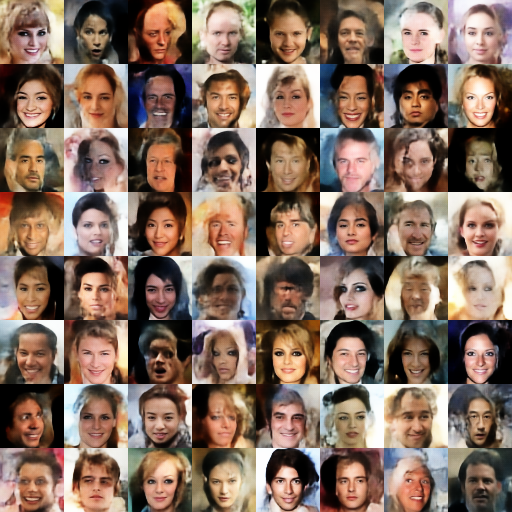} \\
	\vspace{2mm}
	\includegraphics[width=0.42\textwidth]{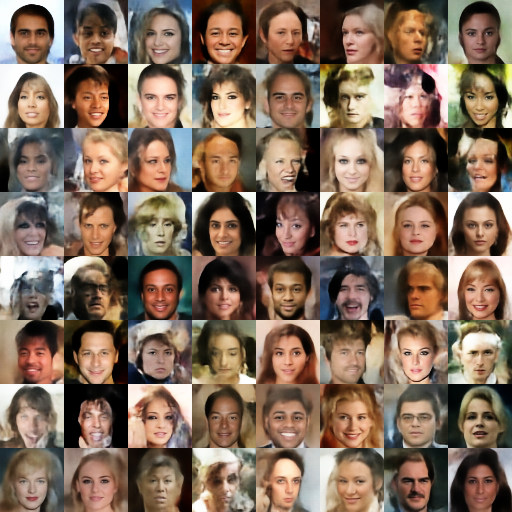} \hspace{2mm}
	\includegraphics[width=0.42\textwidth]{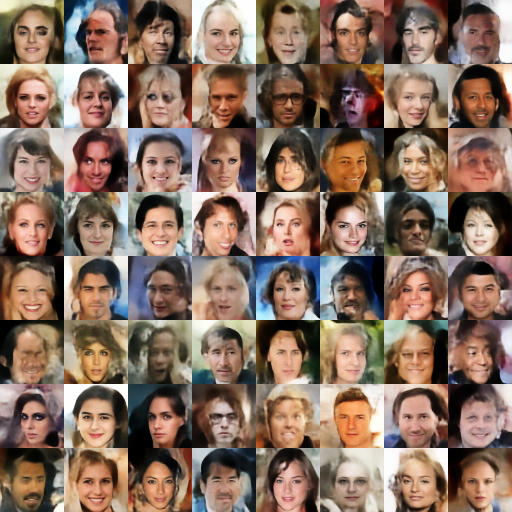}
	\vspace{2mm}
	\includegraphics[width=0.42\textwidth]{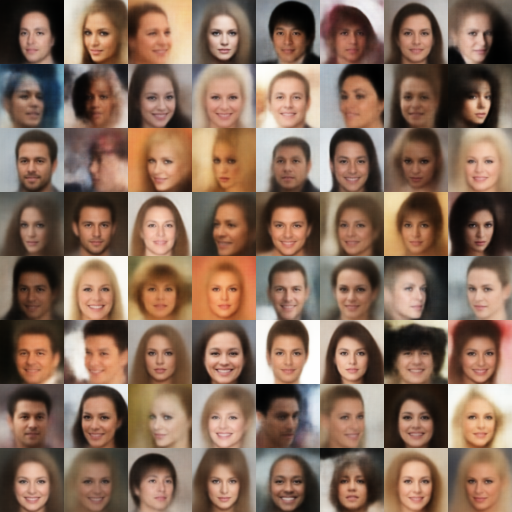} \hspace{2mm}
	\includegraphics[width=0.42\textwidth]{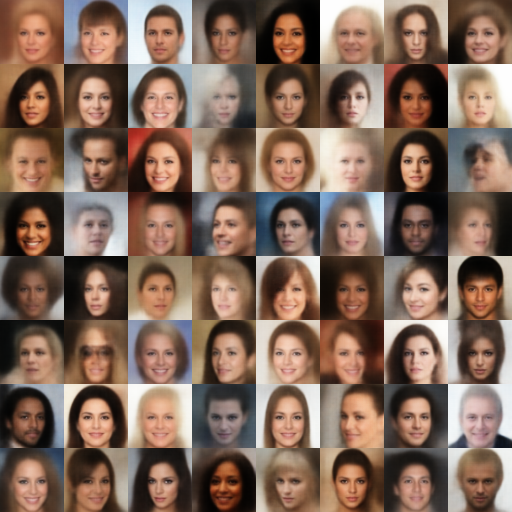}
	\caption{Random CelebA Samples: \emph{Top row:} \proposed, \emph{Middle row:} Autoencoder, \emph{Bottom row:} VAE. \emph{Left:} samples from post-fit GMM prior with 10 components, \emph{Right:} samples from post-fit Gaussian prior.}
	\label{fig:samples_celeba}
\end{figure*}

\noindent {\bf Baseline models.~} Our final objective, although obtained by developing an injective flow and lower bounding its log likelihood, has resemblance with recently proposed regularized autoencoders \citep{ghosh2019variational} which arise as natural models for comparison.

We consider several smoothness regularizers in our evaluations, some of which have also been used by \citet{ghosh2019variational}:

{\bf (i) AE}: Vanilla autoencoder trained with $\ell_2$ reconstruction loss. {\bf (ii) AE+L2}: Autoencoder with an additional $\ell_2$-norm penalty on the decoder parameters (weight decay).
{\bf (iii) AE+SN}: Autoencoder with an additional spectral normalization on each individual layer of the decoder (\ie normalizing the top singular value to be 1), motivated by \citep{miyato2018spectral}.
{\bf (iv) CAE}: We also use contractive autoencoder \citep{rifai2011manifold} as a baseline which penalizes the Frobenius norm of the \emph{encoder's Jacobians}. We use a similar Hutchinson trace stochastic approximation (as used for our objectives) to optimize the Frobenius norm term in the CAE objective. 
\citet{ghosh2019variational} also consider a \emph{gradient penalty regularized AE} which penalizes the Frobenius norm of the decoder's Jacobian, a term which is also present in our objective \eqref{eq:injlower2}. %
We also compare with {\bf (v) VAE} \citep{kingma2013auto} and {\bf (vi) $\bm{\beta}$-VAE} \citep{higgins2016beta}, both with a Gaussian observation model at the decoder's output $N(0,\sigma^2 I_D)$. For VAE, $\sigma^2$ is taken to be 1,  while for $\beta$-VAE, varying $\sigma$ directly controls $\beta$, with $\beta=\sigma^2$. We do not report a comparison with Wasserstein Autoencoders (WAE) as \citet{ghosh2019variational} have shown that the tractable WAE-MMD version is outperformed by regularized autoencoders. %

{\bf Architectures.~} We use convolutional neural net based architecture for both encoder and decoder, each having five layers of convolutions or transposed convolutions, respectively. Strides and kernel-size in the convolutional filters differ across datasets, but stay same for all the models for a given dataset. 
We use a slightly larger network  than \citep{ghosh2019variational} (5 vs. 4 layers), and thus rerun all baseline methods so that the results are comparable. This also results in improved scores for baselines over those reported in \citep{ghosh2019variational}.
We use \emph{elu} activation in both encoder and decoder, and also use batch normalization. %
Latent dimensionality is taken to be $128$ for CIFAR-10 and CelebA, and $32$ for MNIST. More details on the architectures used are provided in the supplementary material. 

\noindent {\bf Hyperparameters and training.~}
Our $\log$-Frobenius norm objective \eqref{eq:final} (referred as \proposedln) has four hyperparameters: (i) variance $\sigma^2$ of the isotropic Gaussian distribution on latent space, which determines the weight on the term penalizing the norm of the encodings $\lVert h(x)\rVert_2^2$, (ii) penalty coefficient $\mu$ on the reconstruction loss, (iii) penalty coefficient $\mu_{in}$ on the injectivity loss term, and (iv) singular value threshold used in the injectivity term $\eta$. We use $\eta=0.1$ in all our experiments. Both penalty coefficients $\mu$ and $\mu_{in}$ are initialized to be $1$ at the beginning of optimization and are increased with each minibatch iteration $i$ as $1+\frac{i^{\nu}}{1000}$, where $\nu$ is searched over $\{1, 1.3\}$. The weight on the prior term $\lVert h(x)\rVert^2$ is searched over $\{0, 10^{-4}, 10^{-3}, 10^{-2}, 10^{-1}, 1\}$.
Our objective \eqref{eq:final2} (referred as \proposed) has an additional hyperparameter $\lambda$ that determines the weight on Frobenius norm regularization term, which we fix to $1$ in all our experiments. As discussed earlier, this will result in a tight bound only for the case when all Jacobian singular values are one. 

For AE+L2, the hyperparameter for the $\ell_2$ regularization term is searched over the set $\{0.001,$ $ 0.01, 0.1, 0.5, 1\}$. For $\beta$-VAE, we search over the standard deviation $\sigma$ of the decoder's distribution (which is related to $\beta$ as $\beta=\sigma^2$ for the Gaussian observation model) over the set $\{0.001, 0.01, 0.05, 0.1, 0.5\}$ (for VAE, $\sigma^2=\beta=1$). For CAE \citep{rifai2011contractive}, the hyperparameter penalizing the encoder's Jacobian norm is searched over $\{0.001,0.01,0.1, 0.5, 1., 10.\}$. 
We train all models using Adam optimizer \citep{kingma2014adam} with batch size of $128$ and a fixed learning rate of $0.001$. All models are trained for $100$k minibatch iterations.  

\noindent {\bf Evaluation.~} 
Evaluation of sample quality is a challenging task \citep{theis2015note}, and several metrics have been proposed in literature for this \citep{salimans2016improved,heusel2017gans,sajjadi2018assessing}. We use the FID score \citep{heusel2017gans} as the quantitative metric for our evaluations, which is one of the most popular metrics and has been used in several recent works
\citep{tolstikhin2017wasserstein,dai2019diagnosing,ghosh2019variational} for evaluating generative models. As discussed earlier, we fit a Gaussian and a mixture of 10 Gaussians on the encoded training data, and use these as prior latent distributions to sample from the model. The covariance matrices for the Gaussian as well as for all mixture components in the GMM are taken to be full matrices. We also report FID scores on the test reconstructions apart from qualitative visualization of reconstructions and samples.

For all models, we report the best FID score obtained using decoder sampling from a post-fit GMM in the latent space. We then report all the other scores (i.e., scores for samples from post-fit Gaussian and test reconstructions) for the same model that gives the best GMM samples FID score. This enables us to assess models in term of their best possible sample generation ability. The proposed injective flow models yield better FID scores than all the baseline models for CelebA and CIFAR10, and are competitive on MNIST where they are outperformed by $\beta$-VAE. For most cases, FID scores for samples with post-fit GMM are better than samples with post-fit Gaussian, except for VAE which we suspect could be due to a convergence issue with GMM fitting. In most cases, \proposedln~yields better FID scores than \proposed, which is expected as \proposedln~uses optimal value of $\lambda=\frac{1}{d}\sum_i s_i^2$ (inequality \eqref{eq:injlower}) as opposed to fixed and likely suboptimal value of $\lambda=1$ used for \proposed~in our experiments.

Randomly generated CelebA samples for autoencoder, VAE, and the proposed model (\proposed) are visualized in Fig.~\ref{fig:samples_celeba}. While VAE samples are blurry and tend to lose fine details, they retain global coherence. On the other hand, samples from \proposed~and autoencoder are sharper with more fine details but also have  undesired visual artifacts in some cases. 
Fig.~\ref{fig:recon_celeba} shows reconstructions of randomly sampled test examples using \proposed, autoencoder and VAE. \proposed~reconstructions preserve more fine details than both autoencoder and VAE (\eg, hair strand for image in the third column), as also reflected by improved FID scores. More generated samples are shown in the supplementary material. 
It should be noted that better sample quality can be achieved by fitting a more expressive prior such as a GMM with more components, a VAE or a flow prior \citep{dinh2016density,papamakarios2017masked}, however care must be taken to not overfit the latent encodings of the training points. 
In principle, a model that can produce good quality test reconstructions has the ability to generate good quality novel samples and the challenge lies in fitting a prior distribution that generalizes well.

\vspace{-3mm}
\section{Discussion}
\label{sec:discuss}
\vspace{-2mm}
We proposed a probability flow based generative model that leverages an injective generator mapping, relaxing the bijectivity requirement. We use a change of variables formula to derive an optimization objective for learning the generator and encoder, where a smoothness regularizer on the generator naturally arises from the probability flow, along with some additional penalty terms. %
This nicely motivates several autoencoder regularizers that have been used in the past, such as in \citet{ghosh2019variational}. 
The proposed model also improves over several regularizers studied in  \citet{ghosh2019variational} in terms of FID scores. %

Relaxing the bijectivity constraint loses many nice properties of invertible flow based generative models, such as tractable likelihood and inference for unseen data. A possible approach to recover these aspects could be to do define a background probability model over the full ambient space $X$ and work with a mixture of foreground distribution over $g(Z)$ coming from probability flow and the background distribution. Investigation of this will be an interesting future direction.  

To enable tractable and efficient training of Injective Flow models, we relied on  lower bounds and stochastic approximation for the Jacobian term, and an amortized encoder trained with penalty method. Future work should investigate the degree to which these approximations are accurate, and whether there are better and more efficient approaches for ensuring invertibility ($g(h(x))=x$) on training points (\eg, augmented Lagrangian methods \citep{bertsekasnonlinear} which have been used in \cite{rezende2018taming}).
A benefit of Injective Flow models is the ability to scale to larger input dimensions. In future work, we plan to improve sample quality and cater to higher-resolution images by scaling the models and fitting more expressive priors.

\bibliography{ml}
\bibliographystyle{icml2019}
\appendix
\onecolumn
\section{Architectures}
We used a similar architecture for all datasets, with 5 convolution layers followed by a dense layer projecting to a mean embedding.

Our architecture resembles that of \citet{ghosh2019variational} but with an additional layer, ELU instead of ReLU nonlinearities, and larger latent dimensions. We list Conv (convoutional) and ConvT (transposed convolution) layers with their number of filters, kernel size, and stride.

\begin{table*}[h]
\begin{small}
    \begin{tabular}{p{0.3\linewidth} p{0.3\linewidth} p{0.3\linewidth}}
    	\toprule
        MNIST & CIFAR-10 & CelebA \\
		\midrule
        
        \multicolumn{3}{c}{Encoder} \\ \midrule

        $x \in \mathcal{R}^{28{\times}28}$ \newline
        $\rightarrow \text{Conv}_{64,4,1} \rightarrow \text{BN} \rightarrow \text{ELU}$\newline
        $\quad \rightarrow \text{Conv}_{128,4,2}\rightarrow \text{BN} \rightarrow \text{ELU}$\newline
        $\quad \rightarrow \text{Conv}_{256,4,2}\rightarrow \text{BN} \rightarrow \text{ELU}$\newline
        $\quad \rightarrow \text{Conv}_{512,4,2}\rightarrow \text{BN} \rightarrow \text{ELU}$\newline
        $\quad \rightarrow \text{Conv}_{512,4,1}\rightarrow \text{BN} \rightarrow \text{ELU}$\newline
        $\quad \rightarrow \text{Flatten} \rightarrow \text{FC}_{32}$
      & $x \in \mathcal{R}^{32{\times}32}$ \newline
        $\rightarrow \text{Conv}_{128,4,1} \rightarrow \text{BN} \rightarrow \text{ELU}$\newline
        $\quad \rightarrow \text{Conv}_{256,4,2}\rightarrow \text{BN} \rightarrow \text{ELU}$\newline
        $\quad \rightarrow \text{Conv}_{512,4,2}\rightarrow \text{BN} \rightarrow \text{ELU}$\newline
        $\quad \rightarrow \text{Conv}_{1024,4,2}\rightarrow \text{BN} \rightarrow \text{ELU}$\newline
        $\quad \rightarrow \text{Conv}_{1024,4,1}\rightarrow \text{BN} \rightarrow \text{ELU}$\newline
        $\quad \rightarrow \text{Flatten} \rightarrow \text{FC}_{128}$
      & $x \in \mathcal{R}^{64{\times}64}$ \newline
        $\rightarrow \text{Conv}_{128,5,1} \rightarrow \text{BN} \rightarrow \text{ELU}$\newline
        $\quad \rightarrow \text{Conv}_{256,5,2}\rightarrow \text{BN} \rightarrow \text{ELU}$\newline
        $\quad \rightarrow \text{Conv}_{512,5,2}\rightarrow \text{BN} \rightarrow \text{ELU}$\newline
        $\quad \rightarrow \text{Conv}_{1024,5,2}\rightarrow \text{BN} \rightarrow \text{ELU}$\newline
        $\quad \rightarrow \text{Conv}_{1024,5,2}\rightarrow \text{BN} \rightarrow \text{ELU}$\newline
        $\quad \rightarrow \text{Flatten} \rightarrow \text{FC}_{128}$\\

\midrule
\multicolumn{3}{c}{Decoder} \\ 
\midrule
        $z \in \mathcal{R}^{32} \rightarrow \text{FC}_{7{\times}7{\times}256}$\newline
        $\rightarrow \text{BN} \rightarrow \text{ELU}$\newline
        $\rightarrow \text{ConvT}_{512,4,1}\rightarrow \text{BN} \rightarrow \text{ELU}$\newline
        $\rightarrow \text{ConvT}_{256,4,1}\rightarrow \text{BN} \rightarrow \text{ELU}$\newline
        $\rightarrow \text{ConvT}_{128,4,2}\rightarrow \text{BN} \rightarrow \text{ELU}$\newline
        $\rightarrow \text{ConvT}_{64,4,2}\rightarrow \text{ELU}$\newline
        $\rightarrow \text{Conv}_{1,4,1}\rightarrow \text{Sigmoid}$
      & $z \in \mathcal{R}^{128} \rightarrow \text{FC}_{8{\times}8{\times}512}$\newline
        $\rightarrow \text{BN} \rightarrow \text{ELU}$\newline
        $\rightarrow \text{ConvT}_{1024,4,1} {\tiny\rightarrow \text{BN} \rightarrow \text{ELU}}$\newline
        $\rightarrow \text{ConvT}_{512,4,2} \rightarrow \text{BN} \rightarrow \text{ELU}$\newline
        $\rightarrow \text{ConvT}_{256,4,2}\rightarrow \text{BN} \rightarrow \text{ELU}$\newline
        $\rightarrow \text{ConvT}_{128,4,2}\rightarrow \ \text{ELU}$\newline
        $\rightarrow \text{Conv}_{3,1,1} \rightarrow \text{Sigmoid}$
      & $z \in \mathcal{R}^{128} \rightarrow \text{FC}_{16{\times}16{\times}512}$\newline
        $\rightarrow \text{BN} \rightarrow \text{ELU}$\newline
        $\rightarrow \text{ConvT}_{1024,5,1} {\tiny \rightarrow \text{BN} \rightarrow \text{ELU}}$\newline
        $\rightarrow \text{ConvT}_{512,5,2}\rightarrow \text{BN} \rightarrow \text{ELU}$\newline
        $\rightarrow \text{ConvT}_{256,5,2}\rightarrow \text{BN} \rightarrow \text{ELU}$\newline
        $\rightarrow \text{ConvT}_{128,5,2}\rightarrow \text{ELU}$\newline
        $\rightarrow \text{Conv}_{3,5,1}\rightarrow \text{Sigmoid}$\\
  \bottomrule
    \end{tabular}
    \end{small}
\end{table*}

\section{Additional Samples}
We visualize additional reconstructed test examples and samples from a post-fit GMM model with 10 mixture components on the latents.
\begin{figure*}[t]
	\centering
	\includegraphics[width=\textwidth]{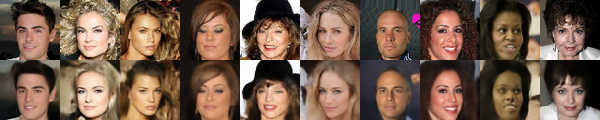}
	\includegraphics[width=\textwidth]{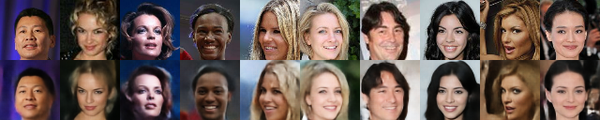}
	\includegraphics[width=\textwidth]{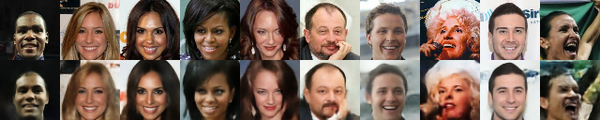}
	\includegraphics[width=\textwidth]{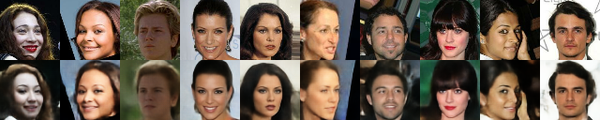}
	\includegraphics[width=\textwidth]{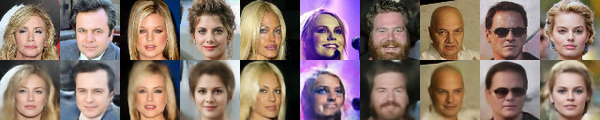}
	\includegraphics[width=\textwidth]{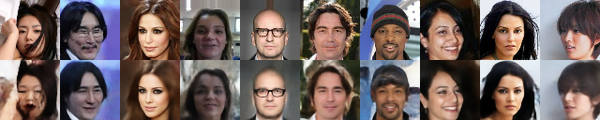}
	\caption{{\bf CelebA} test reconstructions from {\bf \proposed~model}: \emph{Top:} original test image, \emph{Bottom:} reconstructed image.}
	\label{fig:celeba_recon}
\end{figure*}

\begin{figure*}[t]
	\centering
	\includegraphics[width=0.9\textwidth]{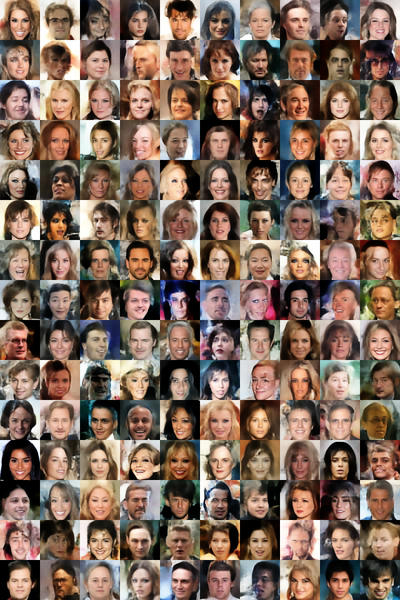} 
	\caption{{\bf CelebA} random samples from {\bf \proposed~model} using the post-fit Gaussian mixture distribution on the latent space.}
	\label{fig:celeba_gmm}
\end{figure*}

\begin{figure*}[t]
	\centering
	\includegraphics[width=\textwidth]{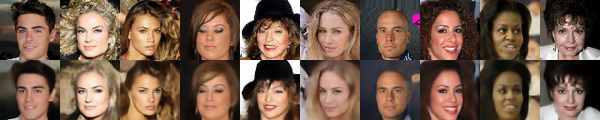}
	\includegraphics[width=\textwidth]{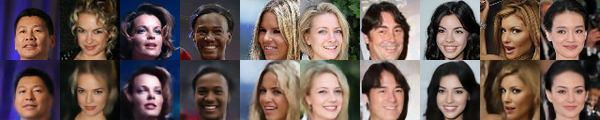}
	\includegraphics[width=\textwidth]{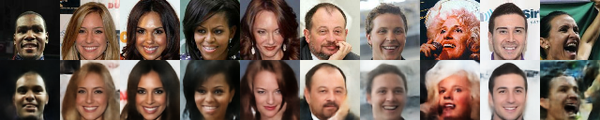}
	\includegraphics[width=\textwidth]{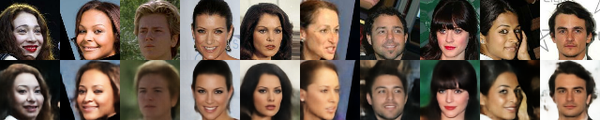}
	\includegraphics[width=\textwidth]{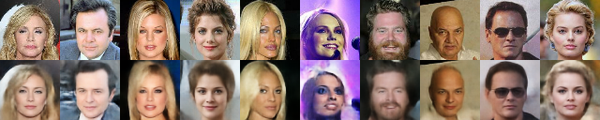}
	\includegraphics[width=\textwidth]{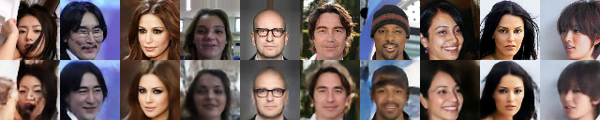}
	\caption{{\bf CelebA} test reconstructions from {\bf Autoencoder}: \emph{Top:} original test image, \emph{Bottom:} reconstructed image.}
	\label{fig:celeba_recon}
\end{figure*}

\begin{figure*}[t]
	\centering
	\includegraphics[width=0.9\textwidth]{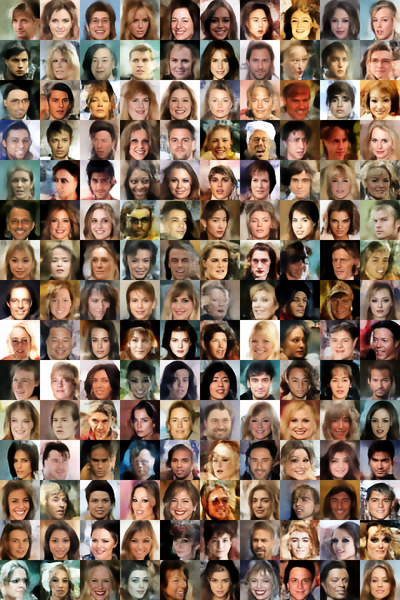} 
	\caption{{\bf CelebA} random samples from {\bf Autoencoder} using the post-fit Gaussian mixture distribution on the latent space.}
	\label{fig:celeba_auto}
\end{figure*}

\begin{figure*}[t]
	\centering
	\includegraphics[width=\textwidth]{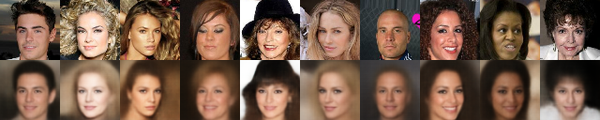}
	\includegraphics[width=\textwidth]{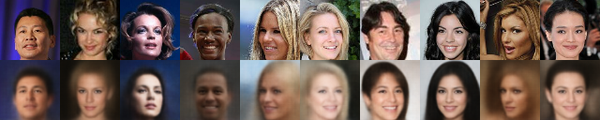}
	\includegraphics[width=\textwidth]{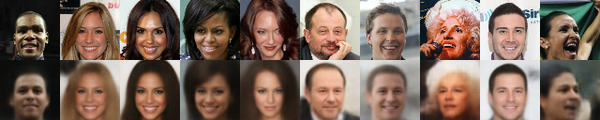}
	\includegraphics[width=\textwidth]{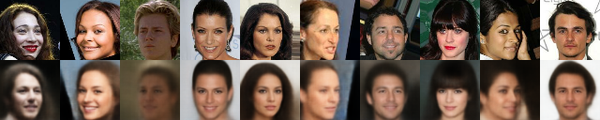}
	\includegraphics[width=\textwidth]{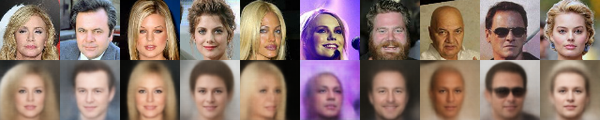}
	\includegraphics[width=\textwidth]{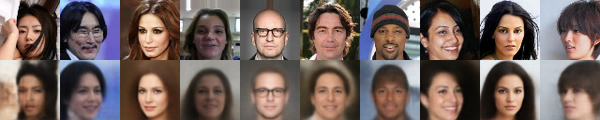}
	\caption{{\bf CelebA} test reconstructions from {\bf VAE}: \emph{Top:} original test image, \emph{Bottom:} reconstructed image.}
	\label{fig:celeba_recon}
\end{figure*}

\begin{figure*}[t]
	\centering
	\includegraphics[width=0.9\textwidth]{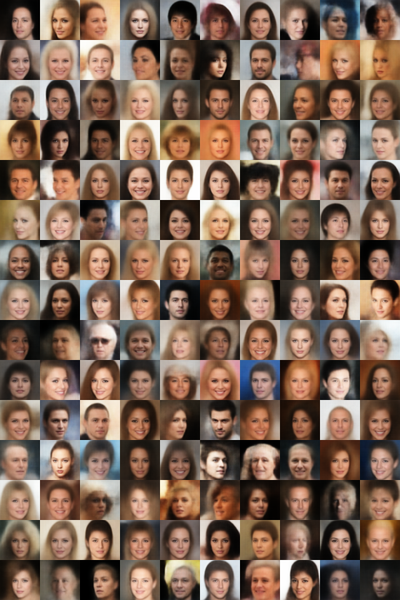} 
	\caption{{\bf CelebA} random samples from {\bf VAE} using the post-fit Gaussian mixture distribution on the latent space.}
	\label{fig:celeba_vae}
\end{figure*}

\begin{figure*}[t]
	\centering
	\includegraphics[width=\textwidth]{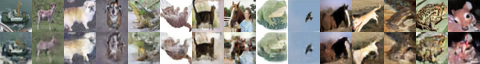}
	\includegraphics[width=\textwidth]{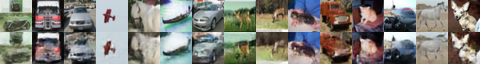}
	\includegraphics[width=\textwidth]{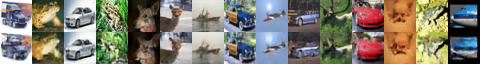}
	\includegraphics[width=\textwidth]{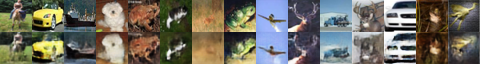}
	\includegraphics[width=\textwidth]{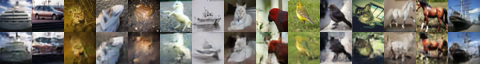}
	\includegraphics[width=\textwidth]{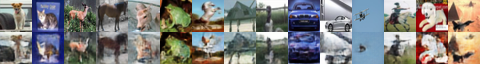}
	\includegraphics[width=\textwidth]{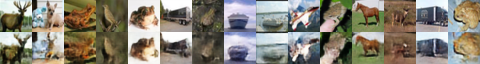}
	\includegraphics[width=\textwidth]{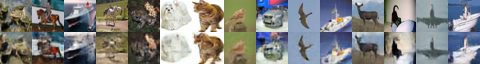}
	\caption{{\bf CIFAR-10} test reconstructions from {\bf \proposed~model}: \emph{Top:} original test image, \emph{Bottom:} reconstructed image.}
	\label{fig:cifar_recon}
\end{figure*}

\begin{figure*}[t]
	\centering
	\includegraphics[scale=1]{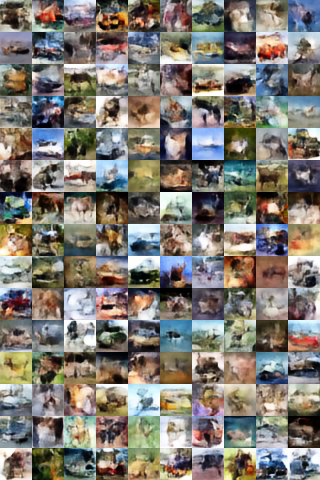} 
	\caption{{\bf CIFAR10} random samples from {\bf \proposed~model} using the post-fit Gaussian mixture distribution on the latent space.}
	\label{fig:cifar_gmm}
\end{figure*}

\begin{figure*}[t]
	\centering
	\includegraphics[width=\textwidth]{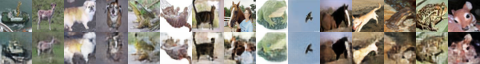}
	\includegraphics[width=\textwidth]{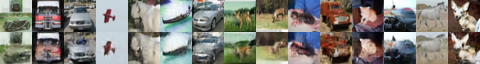}
	\includegraphics[width=\textwidth]{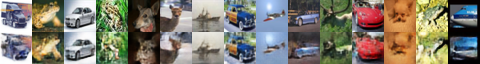}
	\includegraphics[width=\textwidth]{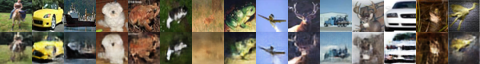}
	\includegraphics[width=\textwidth]{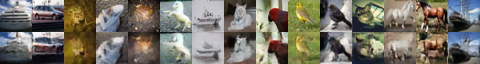}
	\includegraphics[width=\textwidth]{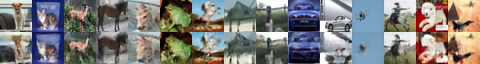}
	\includegraphics[width=\textwidth]{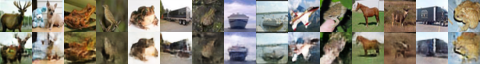}
	\includegraphics[width=\textwidth]{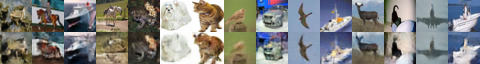}
	\caption{{\bf CIFAR-10} test reconstructions from {\bf Autoencoder}: \emph{Top:} original test image, \emph{Bottom:} reconstructed image.}
	\label{fig:cifar_recon}
\end{figure*}

\begin{figure*}[t]
	\centering
	\includegraphics[scale=1]{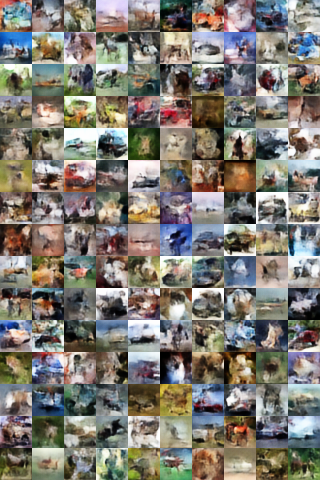} 
	\caption{{\bf CIFAR10} random samples from {\bf Autoencoder} using the post-fit Gaussian mixture distribution on the latent space.}
	\label{fig:cifar_auto}
\end{figure*}

\begin{figure*}[t]
	\centering
	\includegraphics[width=\textwidth]{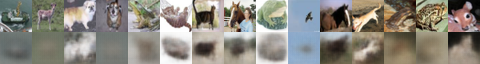}
	\includegraphics[width=\textwidth]{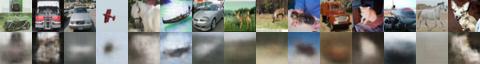}
	\includegraphics[width=\textwidth]{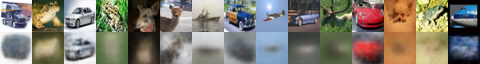}
	\includegraphics[width=\textwidth]{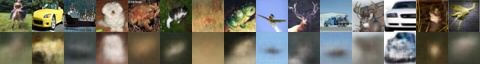}
	\includegraphics[width=\textwidth]{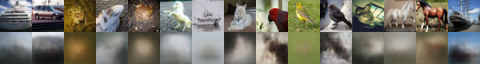}
	\includegraphics[width=\textwidth]{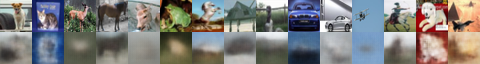}
	\includegraphics[width=\textwidth]{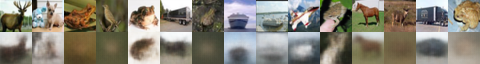}
	\includegraphics[width=\textwidth]{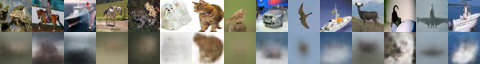}
	\caption{{\bf CIFAR-10} test reconstructions from {\bf VAE}: \emph{Top:} original test image, \emph{Bottom:} reconstructed image.}
	\label{fig:cifar_recon}
\end{figure*}

\begin{figure*}[t]
	\centering
	\includegraphics[scale=1]{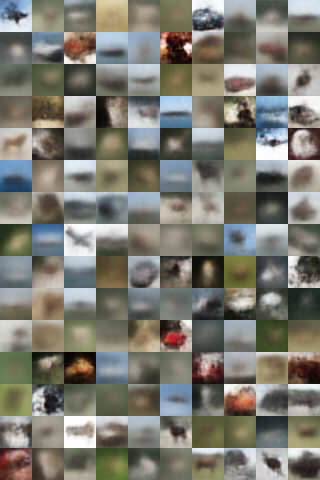} 
	\caption{{\bf CIFAR10} random samples from {\bf VAE} using the post-fit Gaussian mixture distribution on the latent space.}
	\label{fig:cifar_vae}
\end{figure*}

\begin{figure*}[t]
	\centering
	\includegraphics[width=\textwidth]{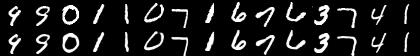}
	\includegraphics[width=\textwidth]{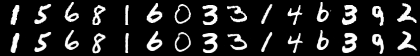}
	\includegraphics[width=\textwidth]{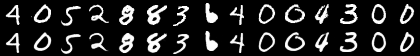}
	\includegraphics[width=\textwidth]{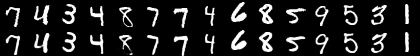}
	\includegraphics[width=\textwidth]{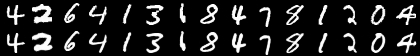}
	\includegraphics[width=\textwidth]{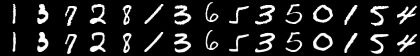}
	\includegraphics[width=\textwidth]{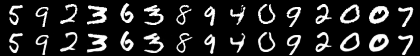}
	\includegraphics[width=\textwidth]{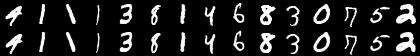}
	\caption{{\bf MNIST} test reconstructions from {\bf \proposed~model}: \emph{Top:} original test image, \emph{Bottom:} reconstructed image.}
	\label{fig:mnist_recon}
\end{figure*}

\begin{figure*}[t]
	\centering
	\includegraphics[scale=1]{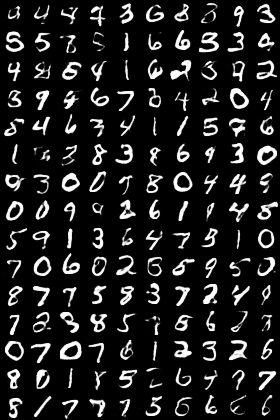} 
	\caption{{\bf MNIST} random samples from {\bf \proposed~model} using the post-fit Gaussian mixture distribution on the latent space.}
	\label{fig:mnist_gmm}
\end{figure*}

\begin{figure*}[t]
	\centering
	\includegraphics[width=\textwidth]{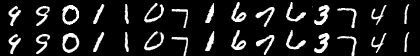}
	\includegraphics[width=\textwidth]{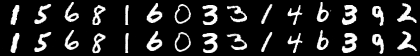}
	\includegraphics[width=\textwidth]{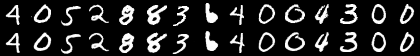}
	\includegraphics[width=\textwidth]{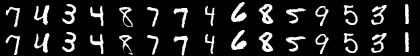}
	\includegraphics[width=\textwidth]{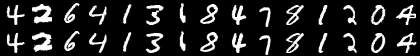}
	\includegraphics[width=\textwidth]{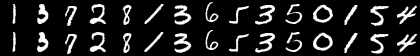}
	\includegraphics[width=\textwidth]{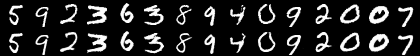}
	\includegraphics[width=\textwidth]{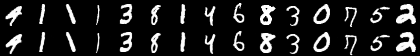}
	\caption{{\bf MNIST} test reconstructions from {\bf Autoencoder}: \emph{Top:} original test image, \emph{Bottom:} reconstructed image.}
	\label{fig:mnist_recon}
\end{figure*}

\begin{figure*}[t]
	\centering
	\includegraphics[scale=1]{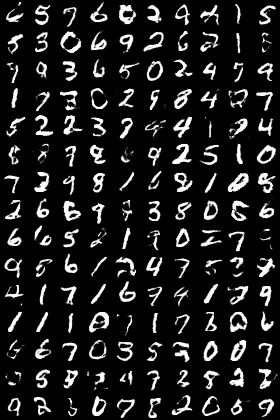} 
	\caption{{\bf MNIST} random samples from {\bf Autoencoder} using the post-fit Gaussian mixture distribution on the latent space.}
	\label{fig:mnist_auto}
\end{figure*}

\begin{figure*}[t]
	\centering
	\includegraphics[width=\textwidth]{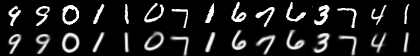}
	\includegraphics[width=\textwidth]{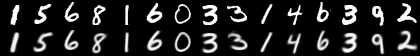}
	\includegraphics[width=\textwidth]{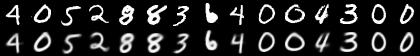}
	\includegraphics[width=\textwidth]{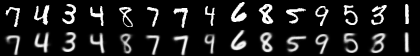}
	\includegraphics[width=\textwidth]{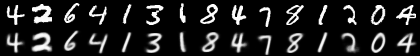}
	\includegraphics[width=\textwidth]{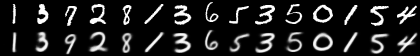}
	\includegraphics[width=\textwidth]{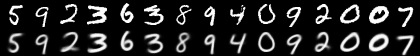}
	\includegraphics[width=\textwidth]{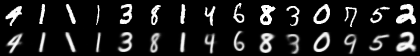}
	\caption{{\bf MNIST} test reconstructions from {\bf VAE}: \emph{Top:} original test image, \emph{Bottom:} reconstructed image.}
	\label{fig:mnist_recon}
\end{figure*}

\begin{figure*}[t]
	\centering
	\includegraphics[scale=1]{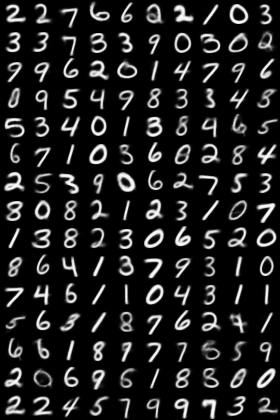} 
	\caption{{\bf MNIST} random samples from {\bf VAE} using the post-fit Gaussian mixture distribution on the latent space.}
	\label{fig:mnist_vae}
\end{figure*}

\end{document}